\documentclass[11pt]{article}
\usepackage[margin=1in]{geometry}
\usepackage{amsmath}
\usepackage{amssymb}
\usepackage{booktabs}
\usepackage{graphicx}
\usepackage{microtype}
\usepackage{url}
\usepackage[hidelinks,breaklinks=true]{hyperref}
\let\oldthebibliography\thebibliography
\renewcommand{\thebibliography}[1]{\oldthebibliography{#1}\sloppy}

\title{When Does Complexity Conditioning Help a Frozen\\
Sentence Embedding? A Controlled Study of Per-Sentence\\
and Pair-Level Difficulty Adaptation}

\author{Suhwan Hwang\thanks{Correspondence: \texttt{hwangsuhwan0109@gmail.com}. Code and logs to accompany the camera-ready.}}
\date{}

\begin{document}
\maketitle

\begin{abstract}
A common intuition is that sentence embeddings should adapt to the difficulty of the input.
We test this intuition in a controlled, multi-seed setting: a lightweight \emph{post-encoder}
adapter attaches to a frozen \texttt{Qwen3-Embedding-0.6B} encoder, accessing only its final
pooled embedding, and is evaluated on four paraphrase and semantic-similarity tasks (PAWS,
MRPC, QQP, STS-B).
The na\"ive form of the idea fails: surface-based per-sentence complexity is nearly
uncorrelated with frozen-baseline error (Pearson $\approx 0.05$) and provides no advantage
over constant or shuffled controls, while degrading a saturated baseline.
Even when the target is aligned to a non-circular pair-difficulty signal, the per-sentence
gate still cannot reliably capture difficulty because difficulty is primarily a property of
the pair, not the individual sentence.
In contrast, a small pair-level residual gated by a held-out cross-encoder difficulty signal
yields consistent gains on the larger and graded tasks, including $+0.022$ Spearman on STS-B
and $+0.037$ on QQP, while remaining anchored to the frozen baseline across all seeds.
Because this useful form operates on sentence pairs rather than individual sentences, the
resulting model is best understood as a lightweight re-ranker over cached frozen embeddings,
not a replacement single-vector embedding; we make no state-of-the-art claim.
Our contribution is a controlled account of when difficulty-aware adaptation helps and when it
fails, together with a pre-training diagnostic that predicts the available headroom.
\end{abstract}

\section{Introduction}

Single-vector sentence embeddings underpin semantic similarity, retrieval, clustering, and
classification \cite{sbert,simcse,mteb}. A recurring intuition is that not all inputs should be
treated identically: longer, syntactically involved, or semantically ambiguous sentences are
``harder,'' and an embedding model might benefit from adapting its geometry to this difficulty.
This paper asks a narrow but practically important question: \emph{does conditioning a frozen
sentence embedding on a measure of input difficulty actually improve it, and if so, under what
conditions?}

We study this with a deliberately conservative design. Rather than fine-tuning the encoder, we
attach a lightweight \emph{post-encoder} adapter that consumes only the baseline's final pooled
embedding $x$. This frozen, embedding-only interface has two advantages. First, it is
maximally portable: it admits any sentence encoder---including ones behind an API---and the
embeddings can be cached once, so the cost of the study is dominated by a single forward pass.
Second, it isolates the contribution of difficulty conditioning from the much larger effect of
end-to-end representation learning, which is what a controlled study requires.

Within this interface we examine a progression of increasingly faithful instantiations of the
difficulty-conditioning idea, each evaluated against tight controls and across multiple seeds:

\begin{enumerate}
\item a \emph{per-sentence} complexity scalar, derived from surface syntax, that conditions an
elementwise scaling of the embedding;
\item the same per-sentence gate, but with the complexity target replaced by a non-circular,
error-aligned hard-pair difficulty signal;
\item a \emph{pair-level} conditioning that preserves the frozen-baseline anchor and adds a
small, difficulty-gated residual to the pair similarity;
\item the pair-level model with a \emph{held-out cross-encoder margin} as the difficulty
target, confirmed across four tasks.
\end{enumerate}

The central methodological device throughout is \emph{signal isolation}: for every claimed
gain we ask whether a control with the difficulty signal removed (constant scale, shuffled
complexity, or an ungated residual of matched capacity) achieves the same effect. This guards
against attributing to ``complexity'' what is in fact extra parameters or a generic residual.

Our findings are summarized as follows. The na\"ive per-sentence form does not work: the
surface complexity signal is essentially uncorrelated (Pearson $\approx 0.05$) with where the
baseline actually errs, and conditioning on it matches a constant-scale control while hurting a
near-saturated baseline. Aligning the target to a non-circular pair-difficulty measure restores
the correlation (up to $0.98$) but still yields no advantage over the constant control, because
hard-pair difficulty is a property of the \emph{pair} and is only weakly recoverable from a
single sentence's frozen embedding. The mechanism becomes effective only when three conditions
hold simultaneously: pair-level conditioning, a small baseline-preserving residual, and a
difficulty target supplied by an independent cross-encoder rather than a surface proxy. Under
these conditions, a cross-encoder-gated pair residual yields consistent, sign-stable gains on
the graded and large tasks (e.g.\ $+0.022$ Spearman on STS-B and $+0.037$ on QQP, with every
seed positive), never falls below the frozen-baseline Spearman in the multi-task confirmation,
and does not degrade retrieval. Because this useful form necessarily operates on sentence
\emph{pairs}, the final model is best understood as a minimal pair-level adapter---a lightweight
re-ranker over cached frozen bi-encoder embeddings---rather than a drop-in replacement for the
single-vector embedding. The study thus begins with an embedding-only question and arrives at a
pair-scoring answer: the negative result rules out pure per-sentence adaptation, and the only
consistently useful positive result lives at the pair-scoring stage, which changes the
deployment interface from single-vector embedding to lightweight re-ranking.

\paragraph{Contributions.}
(1) A controlled, multi-seed study that isolates the conditions under which difficulty
conditioning of a frozen sentence embedding helps, rather than a performance-maximizing system.
(2) A clear negative result---per-sentence complexity scaling does not help even when the
target is well aligned---together with the mechanistic reason (difficulty is a pair property).
(3) A positive result---a small, baseline-preserving, cross-encoder-gated pair residual that
improves the graded/large tasks and never falls below the frozen-baseline Spearman in the
multi-task confirmation, with the explicit caveat that this useful form is pair-level and is
therefore a lightweight re-ranker rather than a pure single-vector embedding---and an analysis
of why the cross-encoder margin generalizes where surface proxies do not. (4) A cheap
pre-training diagnostic, computed from frozen embeddings alone, that predicts the available
headroom and correctly forecasts where adaptation helps or hurts; in practice, before training a
difficulty-aware adapter one should first check whether the proposed difficulty signal correlates
with frozen-baseline error, otherwise the adapter tends to learn a constant or self-suppressing
correction. We make no state-of-the-art claim; the contribution is the controlled understanding
and its reproducibility.

\section{Related Work}

\paragraph{Sentence embeddings.}
Siamese-encoder methods such as Sentence-BERT \cite{sbert} and contrastive methods such as
SimCSE \cite{simcse} produce single-vector representations that are scored by cosine similarity.
The Massive Text Embedding Benchmark \cite{mteb} evaluates such models across retrieval,
similarity, classification, clustering, and reranking. Modern open-weight encoders, including
the Qwen3 embedding family \cite{qwen3emb}, occupy the top of these leaderboards. We treat such
an encoder as a fixed, black-box producer of pooled vectors and study what a thin adapter can
add on top of it without touching its weights.

\paragraph{Bi-encoders versus cross-encoders.}
Bi-encoders embed each input independently, enabling fast nearest-neighbor retrieval but
limiting cross-input interaction; cross-encoders jointly encode a pair and are substantially
more accurate at fine-grained discrimination, at the cost of $O(n^2)$ scoring
\cite{bert_rerank}. A standard pattern is to retrieve with a bi-encoder and re-rank with a
cross-encoder, or to distill the cross-encoder into the bi-encoder \cite{distillation_ce}. Our
positive result sits between these regimes: it keeps the frozen bi-encoder but adds a small
pair-conditioned correction and, crucially, uses a cross-encoder only as a \emph{difficulty
signal} rather than as the scorer.

\paragraph{Difficulty- and hard-negative-aware training.}
The notion that examples differ in difficulty drives hard-negative mining in dense retrieval,
where harder negatives yield stronger encoders \cite{dpr,ance}. These methods change the
training distribution of a model that is being learned end-to-end. We instead ask whether an
explicit, predicted per-input or per-pair difficulty can usefully \emph{condition} a frozen
representation at inference time, and we measure this against controls that remove the
difficulty signal while keeping the added capacity.

\paragraph{Operational complexity targets.}
Surface and syntax-inspired readability features (length, clause counts, subordination markers)
provide cheap, reproducible scalars. We use such a feature engine to define an operational
complexity label, but we do not claim it is a gold linguistic annotation; its role is to serve
as one candidate difficulty signal among several, and our experiments show it is poorly aligned
with embedding error except on adversarial paraphrase data. We contrast it with a held-out
cross-encoder margin, which we find to be a far more robust difficulty signal.

\paragraph{Tasks.}
We evaluate on semantic textual similarity (STS-B \cite{stsb,stsb_dataset}), and three
paraphrase-style tasks: PAWS \cite{paws}, whose adversarial pairs decouple lexical overlap from
meaning, and MRPC \cite{mrpc} and QQP \cite{qqp} from GLUE \cite{glue}. SNLI \cite{snli} is used
only in preliminary pipeline checks. These tasks span a wide range of sentence length, label
granularity (graded vs.\ binary), and baseline saturation, which is exactly the variation our
study requires.

\section{Method}

We describe the frozen-baseline interface and then the four instantiations of difficulty
conditioning that the experiments compare. Throughout, a sentence $a$ has a frozen baseline
embedding $x_a \in \mathbb{R}^{d}$ produced by the baseline encoder, and $\cos(\cdot,\cdot)$
denotes cosine similarity.

\paragraph{Notation.}
A few symbols are shared across arms because the arms share one training objective
(Section~\ref{sec:method-obj}); we make their per-arm meaning explicit here. We write $s$ for the
score produced by the \emph{active} arm: the cosine between adapted vectors for the per-sentence
arms (Section~\ref{sec:method-persentence}), and $s_{\mathrm{ungated}}$ or $s_{\mathrm{gated}}$
for the pair-level arms (Section~\ref{sec:method-pair}). We write $c$ for the predicted
complexity or hardness---$c_a$ for a per-sentence prediction and $c_{ab}$ for a pair-level
prediction---and $\bar c$ for the corresponding supervision target. We use $\hat x_a$ for the
dimension-aligned frozen baseline embedding and $\gamma\in(0,1)$ for the (small) residual scale.

\subsection{Frozen post-encoder interface}

The baseline encoder is called only through its public encode method and is never
differentiated: $x_a$ is computed under \texttt{no\_grad} and detached. A trainable linear map
and layer normalization produce a working representation
\begin{equation}
h_a = \mathrm{LayerNorm}(W x_a + b).
\end{equation}
All trainable parameters live downstream of $x_a$; gradients never reach the encoder. Because
$x_a$ is fixed, embeddings are computed once and cached. This interface requires only a pooled
vector and so admits any encoder, open or closed.

\subsection{Per-sentence complexity scaling}
\label{sec:method-persentence}

The first instantiation follows the literal ``complexity-conditioned'' idea. A small predictor
estimates a scalar complexity from the working representation,
$c_a = \sigma(\mathrm{MLP}(h_a)) \in (0,1)$, supervised by a target $\bar c_a$ via
$\mathrm{MSE}(c_a, \bar c_a)$. The scalar conditions an elementwise scale that preserves the
identity at initialization,
\begin{equation}
\alpha_a = 1 + \alpha_0 \tanh\!\big(g(h_a, c_a)\big), \qquad
u_a = \mathrm{Normalize}(\alpha_a \odot h_a),
\end{equation}
with $\alpha_0$ a small range. The default target $\bar c_a$ is an operational complexity label
$\sum_i w_i \tilde f_i / \sum_i w_i \in [0,1]$ computed from capped, normalized surface features
$\tilde f_i$ (token and character counts, punctuation, conjunctions, subordination markers,
approximate clause count, and a proxy verb/argument count).

\paragraph{Baseline-preserving variants and controls.}
To stay close to the frozen baseline we also consider a residual/interpolation form
$u_a = \mathrm{Normalize}\big((1-\gamma)\,\hat x_a + \gamma\, u^{\mathrm{sem}}_a\big)$ with a
small $\gamma$, where $\hat x_a$ is the (aligned) baseline embedding. The key controls are:
\textbf{baseline-only} ($u_a = \hat x_a$); \textbf{$h$-only} ($\alpha$ from $h_a$ with the
complexity input zeroed); \textbf{fixed/global scalar} (a constant or globally learned $c$);
and \textbf{shuffled complexity} ($c$ permuted across the batch before $\alpha$). The contrast
between a complexity-conditioned scale and a constant scale is our signal-isolation test.

\subsection{Aligned hard-pair difficulty target}
\label{sec:method-aligned}

Surface complexity need not align with where the baseline errs. We therefore define a
non-circular, error-aligned per-pair difficulty from text and gold labels only,
\begin{equation}
\mathrm{hard}_{\mathrm{lex}}(a,b) = \big|\, \mathrm{overlap}(a,b) - y_{ab} \,\big|,
\label{eq:lexhard}
\end{equation}
where $\mathrm{overlap}$ is token Jaccard and $y_{ab}$ is the gold score (the label for binary
tasks). A non-paraphrase with high lexical overlap, or a paraphrase with low overlap, is hard.
This never uses the baseline embedding, so it cannot be circular, yet by construction it tracks
the surface-similarity axis on which bi-encoders fail. We aggregate it to a per-sentence target
(mean over pairs containing the sentence) for the per-sentence gate of
Section~\ref{sec:method-persentence}.

\subsection{Pair-level conditioning}
\label{sec:method-pair}

Because difficulty is a property of the pair, we move the conditioning to the pair while keeping
the frozen-baseline cosine as an anchor. From the two working representations we form the pair
features $\phi_{ab} = [\,|h_a - h_b|,\; h_a \odot h_b\,]$ and a \emph{raw} (gate-free) residual
\begin{equation}
\delta_{\mathrm{raw}}(a,b) = \tanh\!\big(\mathrm{MLP}(\phi_{ab})\big).
\end{equation}
The \textbf{ungated} arm adds this raw residual, scaled by a small $\gamma$, to the frozen
cosine, so the correction is a bounded perturbation of the baseline:
\begin{equation}
s_{\mathrm{ungated}}(a,b) = \underbrace{\cos(x_a, x_b)}_{\text{frozen anchor}}
\;+\; \gamma\,\delta_{\mathrm{raw}}(a,b).
\label{eq:pairscore}
\end{equation}
A separate head predicts a pair hardness
$c_{ab} = \sigma\!\big(\mathrm{MLP}(\phi_{ab})\big) \in (0,1)$, supervised by a difficulty target
(Sections~\ref{sec:method-aligned} and~\ref{sec:method-ce}). The \textbf{gated} arm multiplies
the raw residual by this predicted hardness and \emph{uses the resulting gated residual in place
of the raw one in the score}:
\begin{equation}
\delta_{\mathrm{gated}}(a,b) = c_{ab}\cdot \delta_{\mathrm{raw}}(a,b), \qquad
s_{\mathrm{gated}}(a,b) = \cos(x_a, x_b) + \gamma\,\delta_{\mathrm{gated}}(a,b)
= \cos(x_a, x_b) + \gamma\, c_{ab}\,\delta_{\mathrm{raw}}(a,b),
\label{eq:gated}
\end{equation}
so that easy pairs (low predicted hardness) receive little correction. The two arms use
identically sized heads; the only difference is whether the predicted hardness multiplies the
residual (and is supervised), which makes ``does the difficulty signal help'' directly testable.

\subsection{Held-out cross-encoder difficulty}
\label{sec:method-ce}

Finally we replace the surface proxy of Eq.~\eqref{eq:lexhard} with a held-out cross-encoder
margin. Let $\mathrm{CE}(a,b)\in[0,1]$ be the similarity from an independent cross-encoder. We
define
\begin{equation}
\mathrm{hard}_{\mathrm{CE}}(a,b) = \big|\, \mathrm{CE}(a,b) - \tfrac{1}{2}(\cos(x_a,x_b)+1) \,\big|,
\label{eq:cehard}
\end{equation}
the gap between a strong independent judge and the cheap bi-encoder cosine. This estimates
``where the embedding is likely wrong'' without using gold labels, using a separate model; it
is the difficulty target used in the gated pair scorer of Eqs.~\eqref{eq:pairscore}--\eqref{eq:gated}.

\subsection{Training objective}
\label{sec:method-obj}

Trained models minimize a similarity-regression term on the score, a batchwise logistic rank
term, a difficulty-prediction term (where a hardness head is present), and---for the
per-sentence vector-scaling arms---a vector anchor term:
\begin{equation}
\mathcal{L} = \mathrm{MSE}(s, y) + \lambda_{\mathrm{rank}}\,\mathcal{L}_{\mathrm{rank}}
+ \lambda_{\mathrm{anc}}\,\mathcal{L}_{\mathrm{anchor}}
+ \lambda_{\mathrm{hard}}\,\mathrm{MSE}(c, \bar c).
\label{eq:objective}
\end{equation}

\noindent\textbf{Symbols.} $s$ is the score of the active arm---the cosine between adapted
vectors for the per-sentence arms, and $s_{\mathrm{ungated}}$ or $s_{\mathrm{gated}}$ for the
pair-level arms (Eqs.~\eqref{eq:pairscore}--\eqref{eq:gated}). $y$ is the gold similarity score,
mapped to $\{0,1\}$ for the binary paraphrase tasks (PAWS, MRPC, QQP) and to $[0,1]$ for graded
STS-B. $c$ is the predicted complexity/hardness ($c_a$ for per-sentence arms, $c_{ab}$ for
pair-level arms), and $\bar c$ is the corresponding supervision target: the surface complexity
label $\bar c_a$ of Section~\ref{sec:method-persentence} in the surface experiments, the
aggregated $\mathrm{hard}_{\mathrm{lex}}$ target of Section~\ref{sec:method-aligned} in the
aligned per-sentence experiments, and $\mathrm{hard}_{\mathrm{CE}}(a,b)$ of
Eq.~\eqref{eq:cehard} in the cross-encoder-gated pair experiments. The hardness term is present
only when a hardness head is used (the gated pair arm and the supervised per-sentence arms).

\noindent\textbf{Rank term.} Within each mini-batch, over all ordered pairs $(i,j)$ whose gold
scores satisfy $y_i > y_j$, the rank term applies
$\mathcal{L}_{\mathrm{rank}} = \mathrm{softplus}\!\big(-(s_i - s_j)/\tau\big)$ (averaged over such
pairs, temperature $\tau=0.05$), encouraging pairs with larger gold scores to receive larger
predicted scores.

\noindent\textbf{Anchor term.} For the per-sentence vector-scaling arms,
$\mathcal{L}_{\mathrm{anchor}} = 1 - \cos(u, \hat x)$ averaged over the batch is a \emph{vector}
anchor that keeps each adapted output $u$ close to its frozen baseline embedding $\hat x$. The
pair-level arms carry no explicit anchor term ($\lambda_{\mathrm{anc}}=0$): there, baseline
preservation is structural, coming from the frozen cosine and the small $\gamma$ in
Eqs.~\eqref{eq:pairscore}--\eqref{eq:gated}.

\noindent\textbf{Weights.} $\lambda_{\mathrm{rank}}, \lambda_{\mathrm{anc}}, \lambda_{\mathrm{hard}}$
weight the corresponding terms; we use the values set in the experiment scripts
($\lambda_{\mathrm{rank}}=0.1$ and $\lambda_{\mathrm{hard}}=0.5$ throughout, and
$\lambda_{\mathrm{anc}}=0.05$ for the per-sentence arms). The rank term targets the ordering
metrics used in evaluation; the anchor term and the small $\gamma$ jointly enforce baseline
preservation.

\section{Experimental Setup}

\paragraph{Baseline encoder.}
Unless noted, the frozen baseline is the open-weight \texttt{Qwen3-Embedding-0.6B}
\cite{qwen3emb} ($d=1024$), used through its standard encode interface with $\ell_2$-normalized
outputs. Embeddings are computed once per task and cached; no encoder weights are updated.
Preliminary checks (Section~\ref{sec:results-persentence}) also reference
\texttt{bge-base-en-v1.5} \cite{bge_model_card}.

\paragraph{Tasks and splits.}
We use PAWS (\texttt{labeled\_final}) \cite{paws}, MRPC \cite{mrpc} and QQP \cite{qqp} from
GLUE \cite{glue}, and STS-B \cite{stsb,stsb_dataset}. Binary labels are mapped to scores
$\{0,1\}$ and graded scores to $[0,1]$. The per-sentence and pair-level studies use $2000$
training and $1000$ evaluation pairs; the multi-task confirmation uses up to $4000$ training and
$1500$ evaluation pairs (capped by split size, e.g.\ MRPC validation has $408$ pairs). All pairs
within a task share one cached sentence table.

\paragraph{Cross-encoder.}
The held-out difficulty judge in Eq.~\eqref{eq:cehard} is
\texttt{cross-encoder/stsb-distilroberta-base} \cite{ce_stsb}, a semantic-similarity
cross-encoder; its scores are cached per pair and used only to define the difficulty target,
never as the evaluation scorer.

\paragraph{Training.}
Adapters are trained with AdamW (learning rate $10^{-3}$, weight decay $10^{-2}$), batch size
$64$--$128$, for $400$--$800$ steps depending on the study, with gradient clipping at $1.0$. The
baseline-preserving residual uses $\gamma=0.25$ unless a ceiling regime ($\gamma=2.0$) is
stated. Per-sentence and pair-level studies use seeds $\{0,1,2\}$; the multi-task confirmation
uses five seeds $\{0,1,2,3,4\}$.

\paragraph{Metrics.}
We report Spearman and Pearson correlation between the (pair) score and gold, and two retrieval
proxies on the evaluation pairs---Recall@1 and MRR---computed by ranking each query's gold
partner against all candidates. For pair-level scorers the candidate matrix is scored by the
pair model. All comparisons are \emph{paired by seed}: we report per-seed deltas and whether
every seed agrees in sign.

\paragraph{Headroom diagnostic.}
Before training, we compute from the frozen embeddings alone: the baseline Spearman, the spread
of the per-pair error $|\cos - y|$, and the Pearson correlation between each candidate difficulty
signal and that error. This diagnostic predicts whether adaptation has aligned headroom to
exploit, and we report it alongside the results.

\paragraph{Reproducibility.}
Each study is a self-contained script that caches embeddings and cross-encoder scores, runs all
arms and seeds, and writes per-run metrics and paired deltas. No metric is hand-entered; all
numbers in Section~\ref{sec:results} are read from these logs.

\section{Results}
\label{sec:results}

We present the four studies in the order of Section~\ref{sec:method-persentence}--%
\ref{sec:method-ce}. All correlations are Spearman unless stated; deltas are paired by seed.

\paragraph{Preliminary check.}
Earlier bounded ablations on \texttt{bge-base-en-v1.5} with STS-B (the per-sentence scaling and
small interpolation/metric variants of Section~\ref{sec:method-persentence}) showed only
sub-$0.001$ Spearman movements over a fixed-interpolation reference and no reliable advantage
for sentence-level complexity supervision over $h$-only and fixed/global controls. This
motivated moving to a modern baseline and to tasks with a wider difficulty range.

\subsection{Per-sentence complexity scaling does not help}
\label{sec:results-persentence}

We pair the frozen \texttt{Qwen3-Embedding-0.6B} with PAWS (long, multi-clause, adversarial)
and STS-B (short, near-saturated), $2000/1000$ pairs, $400$ steps, seeds $\{0,1,2\}$. The
pre-training diagnostic already separates the regimes: on PAWS the frozen baseline reaches
Spearman $0.338$ with a large error spread ($\mathrm{std}\,|\cos-y| = 0.456$); on STS-B it
reaches $0.911$ with a small spread ($0.135$). However, the surface complexity signal is
misaligned: the correlation between sentence length and per-sentence baseline error is only
$+0.05$ on both tasks.

Consistent with this, conditioning gives no leverage. The signal-isolation contrast
(complexity-conditioned scale minus constant-$\gamma$ scale) is $-0.0007$ on PAWS, i.e.\ the
complexity signal adds nothing over a constant; on the saturated STS-B every adaptation arm
\emph{reduces} Spearman (adaptive $-0.0019$ vs.\ baseline, and $-0.0073$ Recall@1). Dropping the
baseline and keeping only the learned projection costs $\approx 0.02$ Spearman on both tasks,
confirming that baseline preservation is load-bearing but that the trained head carries no
information the frozen embedding lacks.

\subsection{Aligning the target restores correlation but not gains}

Replacing the surface label with the non-circular hard-pair target of Eq.~\eqref{eq:lexhard}
(aggregated per sentence) raises the correlation with baseline error from $+0.05$ to $+0.98$ on
PAWS---the alignment problem is solved without ever using the baseline embedding. The full model
now improves over the frozen baseline on PAWS by $+0.0024$ Spearman with all three seeds
positive, larger than the surface target. Yet the signal-isolation contrast remains
$\approx 0$: conditioning still matches a constant scale. The reason is structural: the aligned
target's development MSE ($\sim 0.205$) does not fall below its label variance ($\sim 0.14$), so
a single sentence's frozen embedding only weakly predicts hard-pair difficulty. Difficulty is a
property of the \emph{pair}; a per-sentence gate cannot represent it.

\subsection{Pair-level conditioning makes the difficulty signal active}

Moving to the pair scorer of Eq.~\eqref{eq:pairscore} ($2000/1000$ pairs, $600$ steps, seeds
$\{0,1,2\}$) changes the picture (Table~\ref{tab:pairlevel}). In the baseline-preserving regime
($\gamma=0.25$), an \emph{ungated} residual collapses to a rank-invariant shift ($+0.0000$
Spearman), whereas gating the same residual by predicted hardness yields $+0.0103$ Spearman on
PAWS with all seeds positive while largely preserving retrieval (a small Recall@1 change, $0.676$
vs.\ $0.682$). On the saturated STS-B the gated residual self-suppresses to the baseline, while
the ungated residual slightly hurts. At a near-unconstrained budget ($\gamma=2.0$) the raw residual already captures
the Spearman gain ($+0.0105$) but degrades retrieval (Recall@1 $0.639$), and the hardness signal
becomes redundant for Spearman while harming retrieval further. The difficulty signal is thus
useful specifically in the small, baseline-preserving regime---which is the method's premise.

\begin{table}[t]
\centering
\small
\begin{tabular}{lcc}
\toprule
Arm ($\gamma=0.25$) & PAWS Spearman & PAWS Recall@1 \\
\midrule
Frozen baseline & 0.3384 & 0.6820 \\
Pair residual (no hardness) & 0.3384 & 0.6820 \\
Pair residual, hardness-gated & \textbf{0.3487} & 0.6757 \\
\bottomrule
\end{tabular}
\caption{Pair-level conditioning, baseline-preserving regime, mean over three seeds. The
hardness gate is load-bearing: it turns an inert residual into a consistent (all-seeds-positive)
Spearman gain, with only a small Recall@1 change ($0.676$ vs.\ $0.682$).}
\label{tab:pairlevel}
\end{table}

\begin{figure}[t]
\centering
\includegraphics[width=0.52\linewidth]{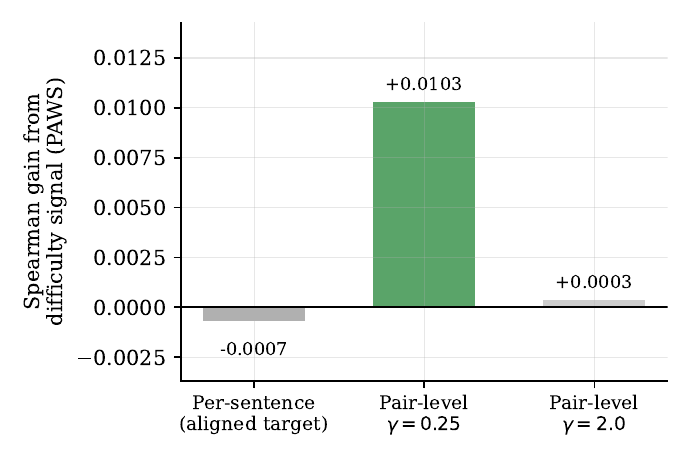}
\caption{Spearman gain attributable to the difficulty signal on PAWS (treatment minus matched
control: complexity-conditioned vs.\ constant scale for the per-sentence stage; hardness-gated
vs.\ ungated residual for the pair-level stages). The signal is active only for the small,
baseline-preserving pair residual.}
\label{fig:isolation}
\end{figure}

Figure~\ref{fig:isolation} summarizes the signal-isolation delta on PAWS across the design
stages: the difficulty signal contributes nothing per-sentence ($-0.0007$) and nothing at the
near-unconstrained budget ($+0.0003$), and becomes active only for the small, baseline-preserving
pair residual ($+0.0103$).

\subsection{A held-out cross-encoder margin confirms across tasks}

\begin{figure}[t]
\centering
\includegraphics[width=0.62\linewidth]{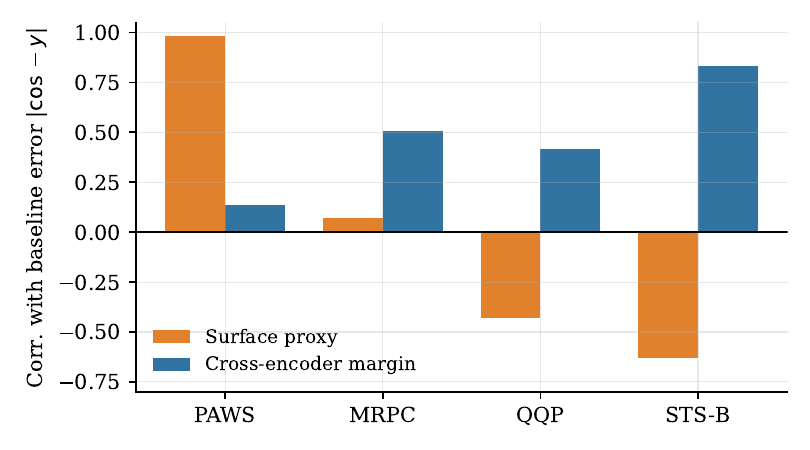}
\caption{Pearson correlation of each candidate difficulty signal with the true per-pair baseline
error $|\cos-y|$ on the evaluation split. The cross-encoder margin is robustly aligned across all
four tasks; the surface proxy is aligned only on PAWS and anti-aligned on QQP and STS-B.}
\label{fig:alignment}
\end{figure}

We replace the surface proxy with the cross-encoder margin of Eq.~\eqref{eq:cehard} and confirm
at multi-task scale (PAWS, MRPC, QQP, STS-B; up to $4000/1500$ pairs; $800$ steps; five seeds;
$\gamma=0.25$). Figure~\ref{fig:alignment} shows that the cross-encoder margin is positively
aligned with the true baseline error on \emph{all four} tasks ($+0.14$, $+0.50$, $+0.42$,
$+0.83$), whereas the surface proxy is aligned only on PAWS (which is built around lexical
overlap, $+0.98$) and \emph{anti}-aligned on QQP ($-0.43$) and STS-B ($-0.63$): the surface
proxy is a PAWS-specific artifact.

\begin{figure}[t]
\centering
\includegraphics[width=0.78\linewidth]{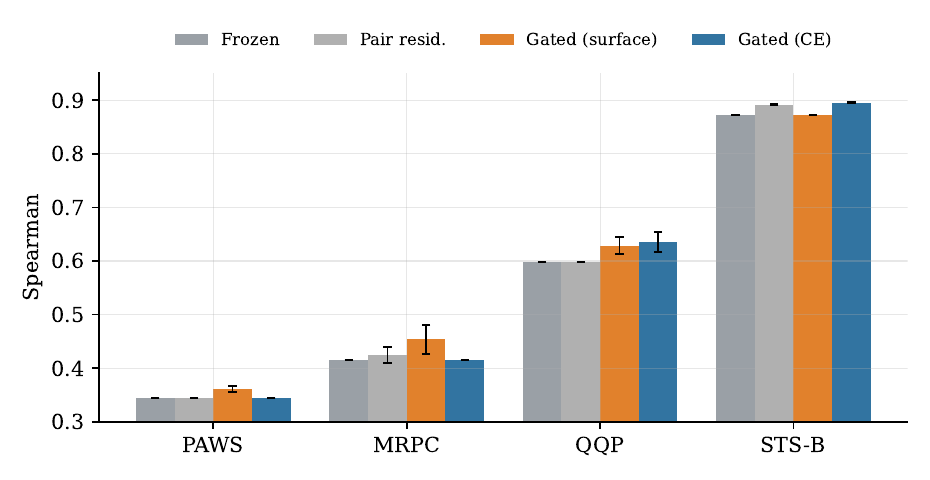}
\caption{Multi-task Spearman by arm (mean over five seeds, $\gamma=0.25$; error bars are one
standard deviation across seeds). The cross-encoder-gated residual never falls below the
frozen-baseline Spearman and improves the graded/large tasks (QQP, STS-B); the surface gate helps
only on its PAWS home turf.}
\label{fig:spearman}
\end{figure}

Table~\ref{tab:crossenc} and Figure~\ref{fig:spearman} report Spearman by arm.\footnote{The
absolute STS-B frozen baseline here ($0.8731$) differs from the $0.911$ of
Section~\ref{sec:results-persentence} only because the two studies score different capped
evaluation subsets of the same frozen cosine---the first $1500$ vs.\ the first $1000$ STS-B
validation pairs. All claims are therefore made through within-study paired comparisons rather
than cross-study absolute values.} The cross-encoder-gated residual never falls below the
frozen-baseline Spearman on any task: it self-suppresses to the baseline on PAWS and MRPC and
improves QQP ($+0.0372$) and STS-B ($+0.0220$, all five seeds positive). It also does not
degrade retrieval: its Recall@1 equals the baseline on PAWS and MRPC and is marginally higher on
QQP ($+0.0012$) and STS-B ($+0.0030$).

We caution against reading Table~\ref{tab:crossenc} by average Spearman alone. The surface-gated
arm attains the highest average, but this is driven almost entirely by PAWS, whose adversarial
construction makes the lexical-overlap proxy naturally aligned with where the baseline errs
(Fig.~\ref{fig:alignment}); off PAWS the surface proxy is unreliable or anti-aligned, the surface
gate degrades MRPC retrieval (from $0.961$ to $0.911$), and it cannot help on the anti-aligned
STS-B and QQP. The appropriate comparison under our diagnostic is thus not the average alone but
cross-task alignment, sign stability, and baseline-preserving behavior. By those criteria the
cross-encoder gate is preferred: it is positively aligned with baseline error on all four tasks,
never falls below the frozen-baseline Spearman, and does not degrade retrieval. We do not claim
it is uniformly best per task---on PAWS the surface proxy is better---only that it is the more
robust and general choice under the paper's diagnostic criterion.

\begin{table}[t]
\centering
\small
\begin{tabular}{lcccc}
\toprule
Task & Frozen & Pair resid. & Gated (surface) & Gated (CE) \\
\midrule
PAWS  & 0.3449 & 0.3449 & \textbf{0.3613} & 0.3449 \\
MRPC  & 0.4156 & 0.4250 & \textbf{0.4537} & 0.4156 \\
QQP   & 0.5988 & 0.5988 & 0.6289 & \textbf{0.6360} \\
STS-B & 0.8731 & 0.8912 & 0.8731 & \textbf{0.8951} \\
\midrule
Average & 0.5581 & 0.5650 & 0.5793 & 0.5729 \\
\bottomrule
\end{tabular}
\caption{Multi-task Spearman by arm (mean over five seeds, $\gamma=0.25$). The cross-encoder
gate never falls below the frozen-baseline Spearman and does not degrade retrieval; the surface
gate attains the highest average but is driven by PAWS and is unreliable off it. ``Pair resid.''
is the ungated residual.}
\label{tab:crossenc}
\end{table}

\section{Discussion}

\paragraph{Three conditions for difficulty conditioning to help.}
Across the five studies the same picture emerges. Difficulty conditioning of a frozen sentence
embedding becomes a real, non-harmful, and \emph{generalizing} effect only when three conditions
hold together: (i) the conditioning is \emph{pair-level}, not per-sentence, because hard-pair
difficulty is not recoverable from a single sentence's frozen embedding; (ii) the correction is
a \emph{small, baseline-preserving residual}, because at large budgets a generic pair residual
already captures the correlation gain and the difficulty signal becomes redundant while harming
retrieval; and (iii) the difficulty target is an \emph{independent cross-encoder margin}, not a
surface proxy or the circular self-error, because only the cross-encoder margin is aligned with
baseline error across tasks. Removing any one condition removes the effect.

\paragraph{Why the cross-encoder margin generalizes.}
A surface proxy encodes a particular hypothesis about difficulty (lexical-overlap mismatch). On
PAWS, which is constructed around that exact phenomenon, the proxy is near-perfectly aligned;
elsewhere it is uninformative or anti-aligned, and gating by an anti-aligned signal can only
hurt or self-suppress. The cross-encoder margin instead measures the gap between a strong
independent judge and the cheap bi-encoder, which is a task-agnostic estimate of ``where the
embedding is wrong.'' This is why it is positively aligned on all four tasks and why the gated
residual it drives never falls below the frozen-baseline Spearman.

\paragraph{A useful safety property.}
The predicted-difficulty gate confers an automatic ``do no harm'' behavior: where the difficulty
signal has no aligned headroom (saturated STS-B for the surface proxy; PAWS/MRPC for the
cross-encoder margin), the gate drives the residual toward zero and the model reverts to the
frozen-baseline score. Combined with the small $\gamma$ and the anchor loss, this means that in
our multi-task confirmation the cross-encoder-gated adapter never reduced the frozen-baseline
Spearman and did not degrade retrieval---an attractive property for a drop-in post-encoder
module, although we verify it only at the bounded scale of this study.

\paragraph{Relationship to re-ranking and distillation.}
Our positive model is a minimal learned re-ranker over frozen bi-encoder embeddings, and it is
worth stating precisely how it differs from two neighbours. Standard cross-encoder re-ranking
uses the cross-encoder \emph{itself} as the inference-time scorer \cite{bert_rerank}; standard
distillation trains a student bi-encoder or scorer to \emph{imitate} the cross-encoder's output
\cite{distillation_ce}. We do neither: the cross-encoder is used only to \emph{define a
difficulty target}---the gap between the cross-encoder similarity and the frozen bi-encoder
cosine (Eq.~\eqref{eq:cehard})---while the inference score remains anchored to the frozen cosine
plus a bounded residual (Eq.~\eqref{eq:pairscore}). The cross-encoder is therefore neither the
scorer nor a target to be imitated; it is an independent estimator of \emph{where} the frozen
embedding needs correction. The method is genuinely related to re-ranking and distillation---it
is a pair-level scorer, not a single-vector embedding---but the experimental question is
different: whether an independently supplied difficulty signal can localize the corrections a
frozen embedding needs. We make no claim that it matches the accuracy of full cross-encoder
re-ranking, which remains an upper reference.

\paragraph{The headroom diagnostic.}
The pre-training diagnostic predicted, from frozen embeddings alone, that STS-B was saturated
(high baseline Spearman, small error spread) and that adaptation would not help there, and that
PAWS had large but structurally specific headroom. These predictions matched the outcomes. We
suggest reporting such a diagnostic before training any post-encoder adapter, as it cheaply
separates ``no headroom'' from ``headroom along an unrecoverable axis.'' Concretely, the
practical takeaway is: before training a difficulty-aware adapter, first check whether the
proposed difficulty signal is correlated with frozen-baseline error; otherwise the adapter is
likely to learn a constant or self-suppressing correction, exactly as we observe for the surface
proxy off PAWS.

\section{Limitations}

The effects we report are modest in magnitude ($\sim 0.02$--$0.04$ Spearman on the graded and
large tasks) and are demonstrated at a bounded scale: a single open-weight bi-encoder, a single
cross-encoder, four English tasks, thousands rather than full-corpus pairs, and a few hundred
optimization steps. We have not tested larger bi-encoders, multilingual data, or full
benchmark-scale evaluation, and we make no state-of-the-art claim.

Two observations caution against over-reading the positive result. First, several
cross-encoder-gated cells only revert to the frozen baseline rather than improving; the gain is
concentrated on the graded/large tasks. Second, on QQP even the anti-aligned surface gate
improved Spearman, which indicates that part of the measured gain comes from generic
pair-residual capacity rather than from correct difficulty ranking. Our signal-isolation
controls bound but do not entirely eliminate this confound.

The positive configuration changes the interface: the model is a pair scorer (re-ranker), not a
single-vector embedding, so it does not directly produce vectors for approximate nearest-neighbor
indices. The difficulty target also depends on the choice and calibration of the cross-encoder;
a poorly matched judge would supply a weaker signal. Finally, the frozen-baseline design isolates
the adapter but does not speak to whether difficulty conditioning would help under full
end-to-end fine-tuning, which is a different and larger question.

\section{Conclusion}

We asked whether conditioning a frozen sentence embedding on input difficulty improves it, and
under what conditions. Through five controlled, multi-seed studies with strict signal-isolation
controls, we showed that the na\"ive per-sentence form does not help---even when its target is
made well aligned with baseline error---because difficulty is a property of the pair that a
per-sentence gate cannot represent. The mechanism becomes effective only when pair-level
conditioning, a small baseline-preserving residual, and a held-out cross-encoder difficulty
target are combined: the resulting gated residual produces consistent, sign-stable gains on
graded and large tasks, never falls below the frozen-baseline Spearman in the multi-task
confirmation, and does not degrade retrieval. We also provide a cheap pre-training diagnostic
that predicts the available headroom from frozen embeddings alone. The gains are modest, and
because the useful form is pair-level the deployment object is a lightweight re-ranker over
cached frozen embeddings rather than a replacement single-vector embedding, so we make no
state-of-the-art claim; the contribution is a precise, reproducible account of when
difficulty-aware adaptation of frozen embeddings is worthwhile. Natural next steps are to scale
the confirmation to larger encoders and full benchmarks, to replace the cross-encoder margin with
a difficulty signal estimable on unlabeled corpora, and to test whether the same three conditions
govern difficulty conditioning under end-to-end fine-tuning.

\bibliographystyle{plain}
\bibliography{references}
\end{document}